\documentclass{article}

\usepackage[square,numbers]{natbib}
\usepackage[final]{ml4mols_2020}

\usepackage[utf8]{inputenc} 
\usepackage[T1]{fontenc}    
\usepackage{hyperref}       
\usepackage{url}            
\usepackage{booktabs}       
\usepackage{amsfonts}       
\usepackage{nicefrac}       
\usepackage{microtype}      
\usepackage[pdftex]{graphicx}

\title{Lipophilicity Prediction with Multitask Learning and Molecular Substructures Representation}

\author{%
  Nina Lukashina\thanks{JetBrains Research. Saint Petersburg, Russia. Corresponding to nina.lukashina@jetbrains.com.} \\
   \And
   Alisa Alenicheva\footnotemark[1] \\
   \AND
   Elizaveta Vlasova\thanks{ITMO University. Saint Petersburg, Russia.} \\
   \And
   Artem Kondiukov\thanks{BIOCAD. Saint Petersburg, Russia.} \\
   \And
   Aigul Khakimova\footnotemark[3] \\
   \And
   Emil Magerramov\footnotemark[3] \\
   \And
   Nikita Churikov\footnotemark[3] \\
   \And
   Aleksei Shpilman\thanks{National Research University Higher School of Economics, Saint Petersburg, Russia.}\hspace{0.1em} \footnotemark[1] \\
}

\begin{document}

\maketitle

\begin{abstract}
Lipophilicity is one of the factors determining the permeability of the cell membrane to a drug molecule. Hence, accurate lipophilicity prediction is an essential step in the development of new drugs. In this paper, we introduce a novel approach to encoding additional graph information by extracting molecular substructures. By adding a set of generalized atomic features of these substructures to an established Direct Message Passing Neural Network (D-MPNN) we were able to achieve a new state-of-the-art result at the task of prediction of two main lipophilicity coefficients, namely logP and logD descriptors. We further improve our approach by employing a multitask approach to predict logP and logD values simultaneously. Additionally, we present a study of the model performance on symmetric and asymmetric molecules, that may yield insight for further research.
\end{abstract}

\section{Introduction}
Molecular property prediction is one of the fundamental issues in chemoinformatics. It is vital for many applications in chemistry and drug discovery \citep{Vamathevan2019, Yang2019}. This paper focuses on the prediction of lipophilicity, a compound's ability to dissolve in fats, oils, and lipids. Lipophilicity is a key physical property for developing small molecule drugs since it reflects the ability of a drug molecule to permeate cell membranes. It is represented by the descriptors logP (the partition coefficient) and logD (the distribution coefficient).

Most of the current state-of-the-art approaches employ the techniques of deep learning. These models solve two sub-tasks: representing a molecule with an appropriate numerical vector and predicting target property for this vector. A combination of trainable molecules' representations with simple regression algorithms as molecular properties predictors demonstrates the most promising results according to the current state of the field.

The standard approach to molecular representations is based on molecular fingerprints, which have limited adaptability to a specific task \citep{Rogers2010}. 

Recently, many promising methods for generating molecule representations have been based on graph representation learning. The growth of the field started with Graph Convolutional Networks \citep{Duvenaud2015, Kearnes2016, Niepert2016, Kipf2016, Wang2019}, which enabled convolutional neural networks to operate directly on graphs. Among other methods, one of the most popular is Message Passing Neural Networks \citep{Gilmer2017, Yang2019}. For example, the JtVAE model \citep{Jin2018} uses the MPNN approach to split the molecular graph into valid chemical substructures. PAGTN model \citep{Chen2019} uses path features in molecular graphs to create global attention layers. OT-GNN \citep{Becigneul2020}, the current state-of-the-art approach for the lipophilicity dataset, computes graph embeddings from optimal transport distances between the set of GNN node embeddings and “prototype” point clouds as free parameters. 

In this paper, we improved the current state-of-the-art result for the prediction of logP/logD properties by introducing a novel approach for substructures representation in molecular graphs and by the application of multitask learning.

\section{StructGNN}

\begin{figure}[h!]
  \centering
  \includegraphics[width=1.0\linewidth]{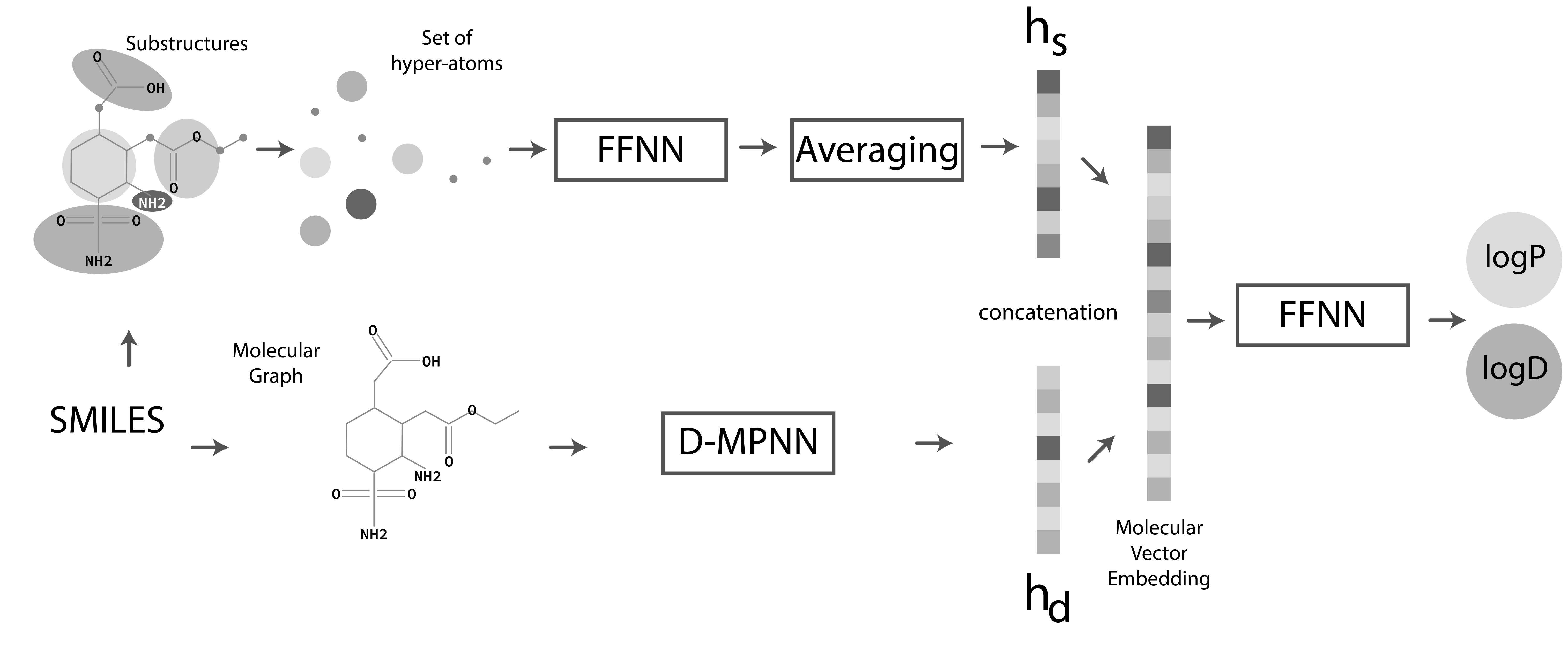}
  \caption{The overall architecture of our StructGNN approach.}
  \label{fig:overall-model}
\end{figure}

Figure~\ref{fig:overall-model} shows the overall network architecture of our method named StructGNN. The model consists of two parallel encoders. The first one is a D-MPNN encoder, which creates a molecular embedding $h_{d}$ through passing edge-focused messages. For a detailed description of the D-MPNN encoder, we refer the reader to the original paper \citep{Yang2019}. 

The second encoder takes in generalized features of extracted substructures and produces embedding $h_{s}$ via the feed-forward neural network (FFNN) and then averaging. We describe this process in detail in Section~\ref{encoder}. 

StructGNN concatenates both embeddings to produce the final molecular vector embedding. This vector could be used with different regression models in order to predict molecular properties. In this paper, we feed it into two layers of fully connected neural network that outputs the final prediction. For the loss function, we use a standard Root Mean Square Error (RMSE) loss.

\subsection{Substructure encoder}\label{encoder}

Substructure encoder considers the molecule's functional groups, namely rings, acids, amines, esters, and sulfonamides. The encoder represents these groups as hyper-atoms. Both hyper-atoms and atoms form a set of nodes for a particular molecule. We approach them as a set rather than a graph because our experiments have shown that introducing connections between nodes does not improve the model's performance. 

We encode each node in the set with a feature vector. We construct features for hyper-atoms by generalizing the features of atoms, originally used in D-MPNN model. Our substructure encoder doesn't create new chemical features, but rather generalizes existing atom's features in order to incorporate the information about the molecule's functional groups. Table~\ref{tab:nodes} shows the list of all used features. 

The feature vector for every node is passed through a one-layer feed-forward neural network. Finally, we average all resulting vectors to produce a single embedding $h_s$ for the whole molecule.

\begin{table}
  \caption{Features for atoms and hyper-atoms}
  \label{tab:nodes}
  \centering
  \begin{tabular}{ll}
    \toprule
    \textbf{Atom's feature}    & \textbf{Hyper-atom's feature}   \\
    \midrule
    Atom type    & Number of each atom type in substructure   \\
    Number of bonds & Number of substructure's \emph{internal} edges \\
    Formal charge   & Substructure formal charge \\
    Chirality   & Number of hydrogen atoms \\
    Number of Hs    & Substructure's aromaticity \\
    Hybridization   &  Sum of atomic masses \\
    Aromaticity     & Substructure's external valence \\
    Atomic mass     & One-hot encoding for the substructure type   \\
    \bottomrule
  \end{tabular}
\end{table}

\subsection{Multitask learning}

We also use multitask learning to predict logP and logD values. Multitask learning aims to improve predictions by solving multiple tasks simultaneously. While logP is used for non-ionizable compounds or the neutral form of ionizable compounds, and logD is used for ionizable compounds, there is an obvious similarity between the two. The Pearson Correlation Coefficient between logP and logD values is 0.66, so we consider descriptors as different targets despite their similarity.

We introduce a new loss function that balances between learning logP and logD properties. If there is only one known value for a molecule (either logP or logD), we calculate the standard RMSE loss function for that values while ignoring the other one. If both values are present, we use mean RMSE between logP and logD errors.

\section{Experiments}

We used a combined dataset of experimental logP and logD values from sources, listed in Table~\ref{tab:datasets}. Data preparation includes SMILES standardization and duplicates removal. The resulting logP and logD datasets contain 13688 and 4166 unique SMILES, respectively. For multitask learning, we merged two datasets and obtained 17603 unique SMILES with logP or logD values. Among them, 251 molecules have both logP and logD values.

\begin{table}
  \caption{Data sources}
  \label{tab:datasets}
  \centering
  \begin{tabular}{lll}
    \toprule
    \textbf{LogP Dataset}     & \textbf{Size (number of samples)}    \\
    \midrule
    PhysProp \cite{physprop} & 13553   \\
    NCI Open Database Compounds \citep{Ihlenfeldt2002}     & 2534   \\
    OChem \citep{Sushko2011}    & 773  \\
    DiverseDataset \citep{Martel2013}     & 707  \\
    \cmidrule(r){2-2}
          &\textbf{17318} in total\\
    \midrule
    \textbf{logD Dataset}     & \textbf{Size (number of samples)}    \\
    \midrule
    lipophilicity \citep{Hersey2015}     & 4200  \\
    \bottomrule
  \end{tabular}
\end{table}

In our experiments, we compare a set of different models to create molecules’ representations for logP/logD prediction. For a fair comparison, we use the same two layers of fully connected neural network as a regression model that takes in that generated representation vector. Our evaluation includes circular fingerprints \cite{Rogers2010} as a baseline method and OT-GNN as the current state-of-the-art approach. We also compare StructGNN with D-MPNN, as it was our base method for improvements, and JtVAE, as it also uses the idea of substructure encoding.

D-MPNN model originally uses an additional vector of 200 global molecular features from RDKit \cite{rdkit}. This approach helps the model to capture global features, while message-passing is fundamentally local in nature \cite{Yang2019}. We followed this approach and used RDKit features in all experiments with D-MPNN and StructGNN models. We used all original features, except MolLogP, as it is calculated logP value from another algorithm. For the StructGNN model, we also excluded structural features, as they are mirrored by generalized hyper-atoms' features. 

We optimized the hyperparameters of every model 4 times on different 25\% validation datasets and then measured the performance on a 20\% test dataset. Table~\ref{tab:results} shows the average RMSE results. Our method achieves new state-of-the-art results in both predicting logP and logD values. 

\begin{table}
  \caption{Results of different models on the lipophilicity descriptors prediction}
  \label{tab:results}
  \centering
  \begin{tabular}{lllll}
    \toprule
    \textbf{Model}    & \textbf{logP (RMSE)}     & \textbf{logP ($R^2$)}     & \textbf{logD (RMSE)}     & \textbf{logD ($R^2$)}  \\
    \midrule
    Morgan Fingerprints   & 0.572+/-0.001  & 0.903+/-0.001  & 0.775+/-0.022   & 0.585+/-0.023 \\
    JtVAE   & 0.511+/-0.009  & 0.922+/-0.003  & 0.640+/-0.016   & 0.712+/-0.015 \\
    OTGNN   & 0.508+/-0.016  & 0.923+/-0.005  & 0.735+/-0.114  & 0.617+/-0.124 \\
    D-MPNN   & 0.464+/-0.002  & 0.936+/-0.001  & 0.601+/-0.008   & 0.749+/-0.007 \\
    StructGNN   & \textbf{0.453+/-0.004}  & \textbf{0.939+/-0.001}  & \textbf{0.591+/-0.014}   & \textbf{0.758+/-0.012} \\
    \midrule
    D-MPNN+Multitask   & 0.442+/-0.004  & 0.944+/-0.001  & 0.554+/-0.004   & 0.805+/-0.003 \\
    StructGNN+Multitask   & \textbf{0.425+/-0.005}  & \textbf{0.948+/-0.001}  & \textbf{0.537+/-0.008}   & \textbf{0.817+/-0.006} \\
    \bottomrule
  \end{tabular}
\end{table}

We also analyzed models' errors depending on the molecule's symmetry. To distinguish between symmetric and asymmetric molecules, we used canonical atom ranking \cite{rdkit}. We've found out that all of the approaches had poor scores for the symmetric compounds, although our model improved the results significantly, as demonstrated in Table~\ref{tab:symmetry}. 

\begin{table}
  \caption{Results of different models on the lipophilicity descriptors predictions for the symmetric and asymmetric molecules}
  \label{tab:symmetry}
  \centering
  \begin{tabular}{lllll}
    \toprule
     \textbf{Model}   & \multicolumn{2}{c}{\textbf{logP (RMSE)}}    &   \multicolumn{2}{c}{\textbf{logD (RMSE)}}       \\
    & \textbf{symmetric}     & \textbf{asymmetric} & \textbf{symmetric} & \textbf{asymmetric}\\
    & \textbf{molecules}     & \textbf{molecules} & \textbf{molecules} & \textbf{molecules}\\
    & \textbf{608 in total}     & \textbf{1380 in total} & \textbf{33 in total} & \textbf{4133 in total}\\
    \midrule
    Morgan Fingerprints   & 0.823+/-0.021  & 0.561+/-0.001  & 1.322+/-0.200   & 0.767+/-0.018 \\
    JtVAE   & 0.677+/-0.048  & 0.504+/-0.008  & 0.953+/-0.186   & 0.636+/-0.014 \\
    OTGNN   & 0.797+/-0.056  & 0.495+/-0.016  & 1.025+/-0.125  & 0.731+/-0.116 \\
    D-MPNN   & 0.722+/-0.063  & \textbf{0.437+/-0.005}  & \textbf{0.846+/-0.133}   & 0.598+/-0.007 \\
    StructGNN   & \textbf{0.671+/-0.017}  & 0.443+/-0.006  & 1.113+/-0.147   & \textbf{0.584+/-0.013} \\
    \midrule
    D-MPNN+Multitask   & \textbf{0.539+/-0.021}  & 0.421+/-0.012  & 0.458+/-0.132   & \textbf{0.536+/-0.016} \\
    StructGNN+Multitask   & 0.541+/-0.005  & \textbf{0.419+/-0.006}  & \textbf{0.367+/-0.113}   & 0.538+/-0.008 \\
    \bottomrule
  \end{tabular}
\end{table}

\section{Conclusion}

We have proposed StructGNN, an extension of the D-MPNN architecture, that encodes additional graph information by extracting molecules' substructures. Our model shows new state-of-the-art results in lipophilicity prediction tasks for logP and logD targets. 

We have further enhanced the performance of the model by applying the multitask approach by training the model to predict logP and logD targets simultaneously. 

Finally, we analyzed predictions for symmetric and asymmetric molecules of all studied models. We show that all models struggle with symmetric molecules and no single model outperforms others both in logP and logD prediction for these molecules. Studies in underlying principles that lead to the disparity between the two can lead to further improvement. 

The source code of our model is available at \url{https://github.com/jbr-ai-labs/lipophilicity-prediction}.

\bibliography{bibl}

\small

\end{document}